\documentclass[conference]{IEEEtran}

\usepackage{booktabs}
\usepackage{amsfonts}
\usepackage{amsmath}
\usepackage{amssymb}
\usepackage{lscape}
\usepackage[normalem]{ulem}
\usepackage{longtable}
\usepackage{xr}
\usepackage{pifont}
\usepackage{multirow}
\usepackage{subcaption}
\usepackage{cite}
\usepackage{algorithmic}
\usepackage{graphicx}
\usepackage{textcomp}
\usepackage{xcolor}
\usepackage{import}
\def\BibTeX{{\rm B\kern-.05em{\sc i\kern-.025em b}\kern-.08em
    T\kern-.1667em\lower.7ex\hbox{E}\kern-.125emX}}
    
\usepackage{tikz}
\usetikzlibrary{tikzmark}
\usetikzlibrary {arrows.meta} 
\usetikzlibrary{positioning}
\usetikzlibrary {shapes.multipart}

\newcommand{\xmark}{\ding{55}}

\useunder{\uline}{\ul}{}

\definecolor{rose}{RGB}{154,26,26}
\definecolor{forest}{RGB}{111,148,96}
\definecolor{deepsea}{RGB}{0,64,128}
\definecolor{powder}{RGB}{184,227,233}
\definecolor{cvprblue}{rgb}{0.21,0.49,0.74}

\def\eg{\emph{e.g}.} 
\def\ie{\emph{i.e}.}

\def\etal{\emph{et al}.}

\usepackage[pagebackref,breaklinks,colorlinks,allcolors=cvprblue]{hyperref}
\usepackage[capitalize]{cleveref}

\crefname{section}{Sec.}{Secs.}
\Crefname{section}{Section}{Sections}
\Crefname{table}{Table}{Tables}
\crefname{table}{Tab.}{Tabs.}

\title{\LARGE \bf 
VLLMs Provide Better Context for Emotion Understanding \\ Through Common Sense Reasoning
}

\begin{document}


\author{
\IEEEauthorblockN{
Alexandros Xenos\textsuperscript{*}, 
Niki M. Foteinopoulou\textsuperscript{*}, 
Ioanna Ntinou\textsuperscript{*}, 
Ioannis Patras, 
Georgios Tzimiropoulos
}
\IEEEauthorblockA{Queen Mary University of London, London, UK\\
\{a.xenos, n.m.foteinopoulou, i.ntinou, i.patras, g.tzimiropoulos\}@qmul.ac.uk}
}

\pagestyle{plain}

\maketitle

\renewcommand\thefootnote{}\footnotetext{\textsuperscript{*}These authors contributed equally to this work}

\begin{abstract}

Recognising emotions in context involves identifying an individual's apparent emotions while considering contextual cues from the surrounding scene. Previous approaches to this task have typically designed explicit scene-encoding architectures or incorporated external scene-related information, such as captions. However, these methods often utilise limited contextual information or rely on intricate training pipelines to decouple noise from relevant information. In this work, we leverage the capabilities of Vision-and-Large-Language Models (VLLMs) to enhance in-context emotion classification in a more straightforward manner.
Our proposed method follows a simple yet effective two-stage approach. First, we prompt VLLMs to generate natural language descriptions of the subject’s apparent emotion in relation to the visual context. Second, the descriptions, along with the visual input, are used to train a transformer-based architecture that fuses text and visual features before the final classification task. This method not only simplifies the training process but also significantly improves performance.
Experimental results demonstrate that the textual descriptions effectively guide the model to constrain the noisy visual input, allowing our fused architecture to outperform individual modalities. Our approach achieves state-of-the-art performance across three datasets, BoLD, EMOTIC, and CAER-S, without bells and whistles. Our code will be made publicly available.

\end{abstract}


\section{Introduction}
\label{sec:intro}
Human emotion recognition is a fundamental task in affective computing with various applications in human-computer interaction (HCI)~\cite{hci}, robotics~\cite{robotics}, and education~\cite{education}. While a considerable amount of research effort has been directed towards discerning apparent emotion from two perspectives: (1) facial expression analysis~\cite{greenaway_context_2018, barrett_context_2011} and (2) body posture and gesture analysis~\cite{nicolaou2011}, these approaches often overlook the crucial contextual information inherent in human emotional expression and perception supported by studies in psychology~\cite{greenaway_context_2018, barrett_context_2011}. The task of in-context emotion recognition has seen some rise in popularity within the affective community; it does, however, remain significantly more challenging than the aforementioned directions.

\begin{figure}
  \begin{center}
    \includegraphics[width=0.5\textwidth]{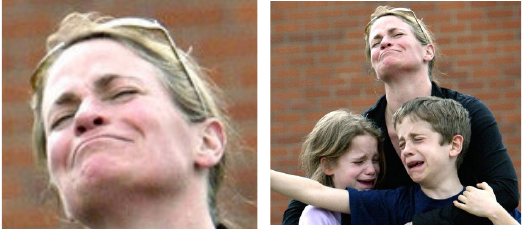}
  \end{center}
    \caption{Example of an ambiguous facial expression from EMOTIC~\cite{kosti2017emotic}. On the left, the apparent emotion of the subject can be confused for discomfort. On the right, the context provides more cues to the ground truth: Pain, Sadness, Suffering}
  \label{fig:example}
\end{figure}
Visual context plays a pivotal role in identifying apparent emotion for both humans and automated systems alike; a facial expression that may indicate surprise in one context might be read as fear in another (\cref{fig:example}). However, while such input contains a lot of information, pixel-level features are typically very noisy and less structured --due to differences in illumination, background clutter, and variations in appearance--, making it harder for automated systems to recognise meaningful patterns with adequate generalisation. Additionally, the visual context can suffer from context bias of the image datasets, where certain visual elements may lead to incorrect or biased interpretations of emotions~\cite{yang_context_2023}.  Inspired by the ability of humans to effectively identify and process only the relevant high-level information in a given visual context~\cite{TREISMAN1980, fei2007we, bisk2020experience}, demonstrated when describing a scene, we propose to transfer the problem into the language space, which is semantically richer and more discrete than the visual space. Specifically, we propose mitigating input noise in the visual input by generating descriptions of the visual context in natural language and using them along with the visual features to train a multi-modal architecture.

Previous works mainly relied on heavy pipelines to effectively recognise emotion while isolating only relevant information from the context. Either using prior knowledge about emotional expression~\cite{mittal_emoticon_2020} or by creating a visual context dictionary~\cite{yang_context_2023,yang2024robust}, these works include limited context information and require more time and resources to train. The main motivation for our work is that language offers a more precise and structured representation of context. By leveraging Vision-and-Language Models (VLLMs) to generate natural language descriptions of images, we utilise VLLMs' inherent common-sense reasoning capabilities. This enables the extraction of relevant contextual information and the filtering out of irrelevant details, mimicking how humans process emotions. More specifically, we propose a two-stage approach; first, we instruct LlaVa~\cite{liu2023llava, liu2023improvedllava}--a state-of-the-art VLLM-- to describe the subject's emotional state relative to the visual context, to bring context into language space. The generated responses contain information regarding emotions (\eg, ``smiling'') and contextual cues (\eg, ``two men discussing'') but contain less noise than pixel-level information. In the second stage, we use the image sample and the generated text descriptions to train a transformer-based architecture that fuses visual and text features before performing the final classification. 
In the ablation studies, we show that the textual descriptions effectively guide the model to constrain the noisy visual input, as the fused architecture performs better than the individual modalities. Finally, we evaluate our work on three popular emotions-in-context datasets and show that our method outperforms previous works and achieves state-of-the-art results for both video and static images without additional training or parameters, thus making our approach effective and novel without bells and whistles. 

Our contributions can be summarised as follows:

\begin{itemize}
    \item We propose a simple two-stage method: first, VLLMs generate natural language descriptions from visual inputs; second, the descriptions and visual features train a multi-modal emotion recognition model. This method prioritises feature quality over parameter quantity.
    \item Extensive experiments show that single modalities result in suboptimal performance, likely due to noise and bias in the image context. However, the language-guided model achieves significantly better results by combining these features. Our ablation studies reveal that including generated contextual descriptions leads to over 10\% mAP improvement on BoLD compared to using visual information alone, with a similar trend on all datasets.
    \item Our method handles both video and static inputs without additional parameters or training, surpassing previous SoTA by 4\% mAP on BoLD and over 2\% accuracy on CAER-S, and SoTA performance on EMOTIC.
\end{itemize}
\section{Related Work}
\subsection{Context Aware Emotion Recognition}

Most approaches in contextual emotion prediction can be split into three main groups: (a) separating the image input into subject and context through cropping and masking, respectively~\cite{ 10.1145/3503161.3547754, 10.1145/3503161.3547755, yang_context_2023, hoang2021context, filntisis_emotion_2020, ruan2020iccv}, (b) extracting features based on prior knowledge regarding emotional expression~\cite{mittal_emoticon_2020, li2021human, yang2022emotion} or (c) examining label relationships~\cite{zhang2019contextaware, li2021human, gao2021graph}. All see an improvement when compared to uni-modal methods, where the input is either just the subject~\cite{zhao2021robust} or the entire image~\cite{jaiswal2020attention}; however, they also face several limitations. More specifically, it may be difficult for Machine Learning (ML) models to identify meaningful patterns when using the visual context without any constraint. Yang~\etal(~\cite{yang_context_2023, yang2024robust}) proposed a causality-based counterfactual de-biasing framework for the CAER task, using a pre-trained causal graph to disentangle context bias. However, such an approach may result in very coarse-grained features that do not adequately represent the visual context. On the other hand, features such as body pose~\cite{mittal_emoticon_2020} may omit interactions of the subject with others and the environment. Finally, Graph Neural Networks (GNN)~\cite{zhang2019contextaware, gao2021graph} and models that rely on label relationships~\cite{li2021human} may result in poor generalisation due to the variable definitions of emotions and dataset bias.

\begin{figure*}[t]
  \begin{center}
    \includegraphics[width=\textwidth]{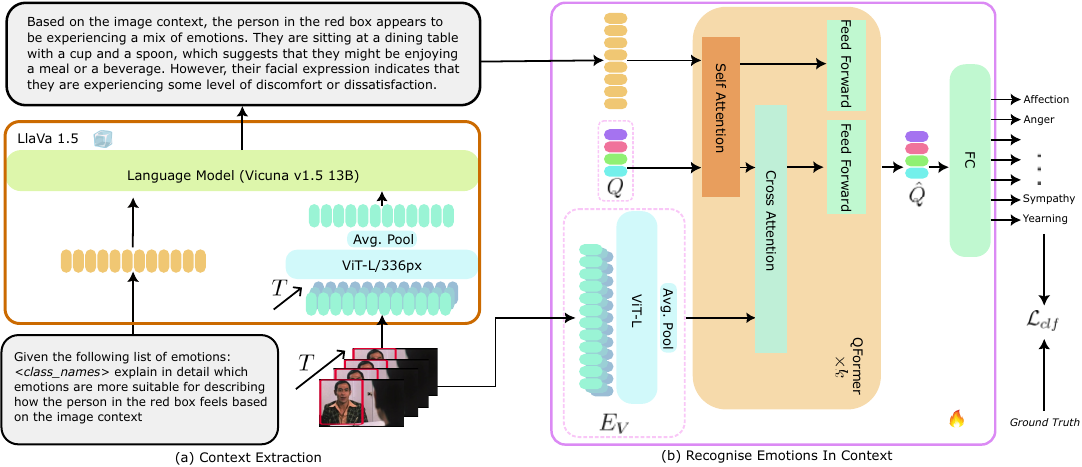}
  \end{center}
    \caption{An overview of our proposed method. We first use LlaVa-1.5~\cite{liu2023improvedllava}, a pre-trained VLLM, to extract language descriptions about the subject's apparent emotion and context (left). The image-description pairs then train our architecture, consisting of a vision encoder ($E_V$), a set of learnable queries ($Q$), a Q-Former module and a Fully Connected layer that performs the final classification on the emotion prediction task. When bounding boxes are available in the annotations, we draw them on the input image so that the model differentiates the subjects and so that the image and generated description are aligned. Our method handles both video and static images with no additional parameters; when video input is used, we average the latent representations of the frame tokens along the temporal dimension.}
  \label{fig:overview}
\end{figure*}

\subsection{VLMs for affective computing}

Several works have attempted to leverage Vision and Language Models for affective prediction tasks. For \textit{Emotion Recognition in Conversations} (ERC), ~\cite{zhang2024dialoguellmcontextemotionknowledgetuned} proposed its reformulation as a retrieval task using LLMs supplemented with two auxiliary tasks for implicit modelling of the dialogue role relationships and future emotional tendencies in conversations. Emotion-LLaMA~\cite{cheng2024emotionllamamultimodalemotionrecognition} uses Multi-modal Large Language Models for \textit{Multi-modal Emotion Recognition} by integrating audio, visual, and text inputs to an LLM using audio and vision encoders. In the aforementioned setting, EmoLLM~\cite{yang2024emollmmultimodalemotionalunderstanding} added an extra module that builds a graph, connecting cluster centres from the generated embeddings. EmoVIT~\cite{Xie_2024_CVPR} proposed a two-stage pipeline where VLLM is used to generate emotion visual instruction data that are in turn used as instructions for Visual Instruction Tuning. For \textit{Facial Expression Recognition}, CLIP-based methods have been proposed by ~\cite{zhao2023prompting, foteinopoulou_emoclip_2023, li2023cliper}. Set-of-Vision prompting (SoV)~\cite{zhang2024visualpromptingllmsenhancing} proposed a multi-stage featutere extraction method where face detection, face numbering, landmark extraction, and spatial relationship analysis are used to assist zero-shot emotion recognition.  For \textit{in-context emotion prediction}, the great body of work is focused on evaluating or studying VLMs' zero-shot and fine-tuning capabilities. EmotionCLIP~\cite{zhang2023learning} investigated the zero-shot generalisation capabilities of a CLIP-based scheme. Behavior-LLaVA~\cite{singh2024llavafindsfreelunch} conducted a large-scale study of user behaviour over image and video content. Given this,  LLaMA-Vid was trained to predict user comments and likes given a media (a video or an image). The model is then fine-tuned for emotion analysis, including in-context emotion prediction. Lei~\etal~\cite{lei2024instructercreformingemotionrecognition} explore chain-of-thought reasoning for CAER, which is a different experimental setting and is not evaluated using the same metrics. Therefore, even though this work showcases some of the VLLM capabilities, it is not comparable to other task-specific models.
In contrast to previous works, our approach is the first to directly tackle in-context emotion prediction using natural language descriptions for implicit context modelling and training a VLM with them.


\section{Methodology}
\label{sec:method}

An overview of our method is shown in~\cref{fig:overview}. We use the pre-trained LlaVa-1.5~\cite{liu2023improvedllava} to generate natural language descriptions of the subject's apparent emotion. Leveraging common-sense reasoning is a key strength of our approach. We then train our architecture --using the visual input and generated descriptions-- which includes a vision encoder ($E_V$), learnable queries ($Q$), and a Q-Former module. The Q-Former performs cross-attention between queries and visual tokens and self-attention between queries and text tokens, with a fully connected layer for final classification. Our method's simplicity, a key contribution of our approach, allows it to handle both videos and static images without additional parameters.

\subsection{Generating Context Descriptions in Natural Language}
\label{sec:desc}

Our proposed method is based upon incorporating context in the form of natural language to recognise emotion in the wild. However, such an annotation effort would be very labour-intensive for human annotators and impractical during inference. For this, we instruct LlaVa to generate sample-level descriptions that describe the apparent emotion of the subject in the given visual context. We draw the bounding box on the frames or images to specify for whom the description should be and prompt the VLLM with:

\texttt{
    USER: $\langle$image$\rangle$
    Given the following list of emotions: {class names}, please explain in detail which emotions are more suitable for describing how the person in the red box feels based on the image context.
}

\noindent where \texttt{\{class names\}} are substituted for each dataset's list of class names. The intuition behind the proposed method is to generate emotionally aware descriptions of the subject and context. That is, rather than generating a generic text description that may be influenced by the input noise (\eg irrelevant background objects), we guide the VLLM to provide descriptions and context relevant to the apparent emotion.
When bounding boxes are not available, the prompt is modified as follows:

\texttt{
    USER: $\langle$image$\rangle$
    Given the following list of emotions: {class names}, please explain in detail which emotions are more suitable for describing how the person feels based on the image context.
}

\begin{figure*}[ht]
  \begin{center}
    \includegraphics[width=0.9\textwidth]{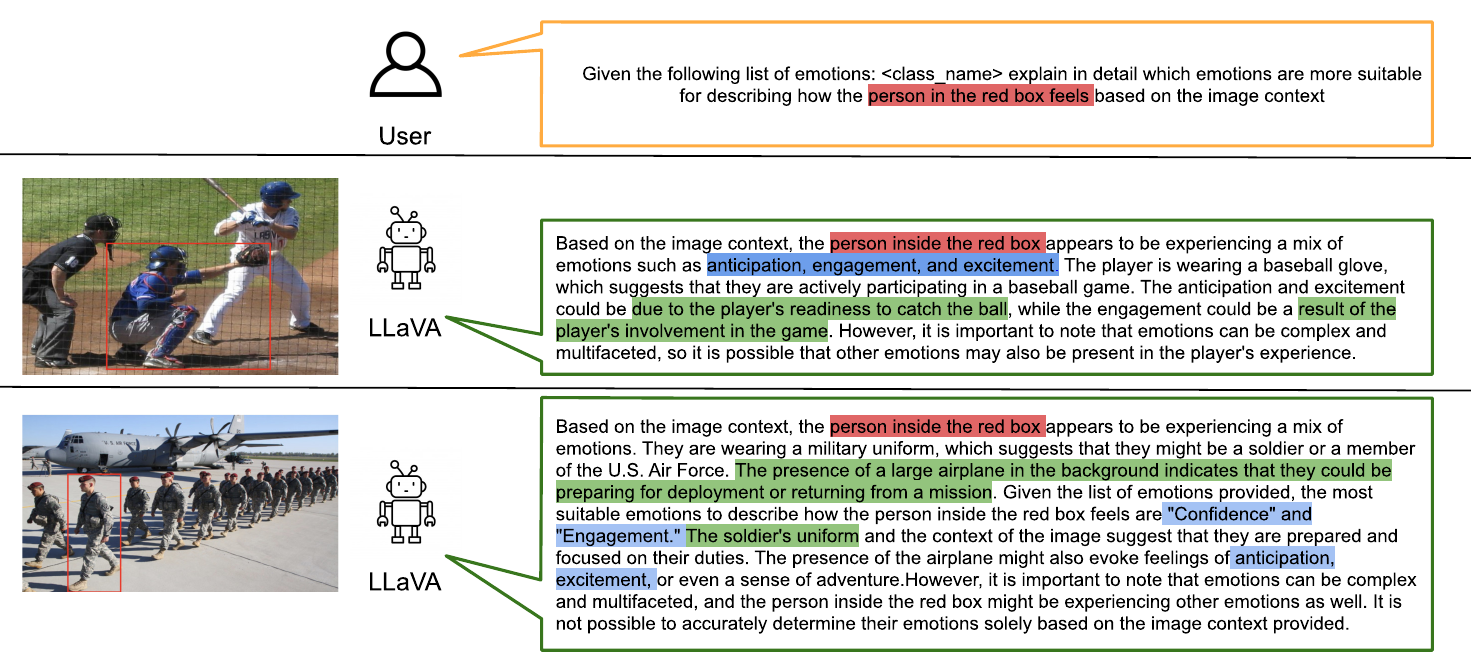}
  \end{center}
    \caption[placeholder]{Examples of contextual descriptions generated for the EMOTIC~\cite{kosti2017emotic} dataset using LLaVa-1.5~\cite{liu2023improvedllava}. The responses are typically person-specific, referring to ``the person inside the red box,'' with guesses regarding the person's apparent emotion (marked with blue) and explanations for these guesses (marked with green).\footnotemark}
  \label{fig:desc}
\end{figure*}

The descriptions often cover details about the emotional state of the subject as well as their interaction with others and the environment. We observe that these include information regarding facial expressions (\eg, ``smiling'') and body pose (\eg, ``pointing'') or more generic elements (\eg ``holding cards'' or ``having a conversation'') but ignore irrelevant details (\eg details of clothing) seen in~\cref{fig:desc}.

The proposed method to extract sample-level descriptions can be adjusted for video input by averaging the latent representations of the video frames. In ~\cref{fig:overview}~(a), we see that LlaVa, similar to most VLLMs, is composed of a Large Language Model (LLM) that generates a response given language tokens (the prompt) and image tokens taken from a vision encoder --in this case, a Vision Transformer architecture (ViT)~\cite{dosovitskiy2020image}. To extract a single description for each video sample, we propose averaging the visual tokens of the frames along the temporal dimension before forwarding them to the LLM. As the LLM module expects a set of spatial representations, we average the representations of the sampled frames for each spatial position.

\subsection{Multi-modal In Context Emotion Prediction} 
\label{sec:architecture}

Our architecture consists of four key components: a vision encoder ($E_V$), a set of learnable queries ($Q$), a Q-Former module~\cite{li_blip-2_2023}, and finally, a fully connected layer that performs the final classification. 

\textbf{Vision Encoder:} The vision encoder consists of a ViT~\cite{dosovitskiy2020image} module used to extract visual features. Given an image $X_v \in \mathbb{R}^{H \times W \times C}$, where $H\times W\times C$ represent the height, width and channels, the image is divided into $P \times P \times C$ non-overlapping flattened patches. 
The vision encoder $E_V$ maps the patches into visual tokens using a linear embedding layer so that each image is represented as a set of $\frac{H}{P} \times \frac{W}{P} \times D$ dimensional tokens, where $D$ is the output dimension of the embedding layer. The visual tokens and a pre-pended learnable classification token are fed into a standard Transformer encoder layer.
The last hidden state of the visual tokens is used as input in the Q-Former. Formally:

\begin{equation}
    \hat{X_v} = E_V(X_v) \in \mathbb{R}^{(\frac{H}{P} \times \frac{W}{P} + 1) \times D}
\end{equation}

If the input is a video, which includes the additional temporal dimension, we add an adaptive average pool along the temporal dimension after the Vision Encoder so that for a video input $V = {X_v^0, \dots, X_v^T}$, where $T$ is the number of frames (shown in ~\cref{fig:overview}~(b)):
\begin{equation}
    \hat{X_v} = \frac{1}{T} \sum_{i=0}^T \hat{X_v^i}
\end{equation}
\textbf{Text Encoder: } To obtain the text features $X_t \in \mathbb{R} ^{L \times d}$, the description generated as described in the previous section is tokenized to obtain $L$ tokens which are then passed to a linear embedding layer to obtain a vector representation $\Vec{t_i} \in \mathbb{R}^d$ representing each (sub)-token

\begin{equation}
    \Vec{t}_i = E_T(t_i),  \Vec{t}_i  \in \mathbb{R} ^d,
\end{equation} 

\noindent where $d$ is the dimension of the text embedding layer and $X_t = \{\Vec{t}_0, \dots, \Vec{t}_L\}$. 

\textbf{Query Tokens: } The model is initialised with a set of learnable query tokens $Q \in \mathbb{R}^{N\times d}$, that match the dimensions of the text embeddings.

\textbf{Q-Former: }
The query tokens and the text features $X_t \in \mathbb{R} ^{L \times d}$ are concatenated to form a single sequence:

\begin{equation}
    Z = Cat[Q;X_t] \in \mathbb{R} ^{(N+L) \times d}
\end{equation} 

This sequence is then fed to the Q-Former module. The Q-Former comprises of $k$ self-attention layers and $k/2$ cross-attention layers (inserted every other transformer block). In each self-attention layer, the queries interact with each other and the description tokens via standard Multiheaded Self-Attention (MSA) \cite{vaswani2017attention}.  The $l$\--th Transformer layer processes the tokens $Z^{l-1} \in \mathbb{R} ^{(N+L) \times d}$ of the previous layer using a series of MSA, Layer
Normalisation (LN), and MLP ($\mathbb{R} ^d$ → $\mathbb{R}^{4d}$ → $\mathbb{R} ^{d}$) layers as follows:

\begin{equation}
    Y^l = MSA(LN(Z^{l-1})) +Z^{l-1},
\end{equation}  
\begin{equation}
    Z^l = MLP(LN(Y^l)) +Y^l
\end{equation}

A single Self-Attention (SA) head is given by
\begin{equation}
    y^l_t = \sum_{t'=0}^{N+L} \sigma\left\{ \frac{(q_t^lk_{t'}^l)}{\sqrt{d_h}}\right\}v_{t'}^l, t=0, ...,N+L,
\end{equation}

\noindent where $\sigma(.)$ is the softmax activation, $q^l_t, k^l_t, v^l_t \in \mathbb{R} ^{d_h}$ are the query, key, and value vectors computed from $z^l_t$ using embedding matrices $W_q,W_k,W_v \in \mathbb{R} ^{d \times d_h}$, $d_h$ is the scale factor in self-attention. Finally, the outputs of the $h$ heads are concatenated and projected using embedding matrix $W_h \in \mathbb{R}^{h_dh \times d}$.
In each cross-attention layer, the queries ($Q$) interact with the image features through  Multiheaded Cross-Attention (MCA) --similar to MSA, but $Q$ are the attention queries and $\hat{X_v}$ are the keys and values.

The attended query vectors $\hat{Q}$ and text tokens are further processed by $2$ small feed-forward networks before the next Q-Former block. 
The output of the Q-Former -- the last state of the attended query vectors -- $\hat{Q} \in \mathbb{R} ^{N \times d}$
is averaged pooled to obtain a single $d$\--dimensional representation and fed to one fully connected layer for the final classification:

\begin{equation}
    \hat{y} = Act(\hat{Q}^TW_{cls}),
\end{equation}

\noindent where $\hat{y} \in \mathbb{R}^C$ is the predicted class vector for $C$ number of classes, $W_{cls}$ is the weight matrix of the prediction layer, and $Act(.)$ is the activation function.
\section{Experimental Results}

\begin{table}[ht!]
\centering
\begin{subtable}[t]{0.5\textwidth}\centering
\small
\setlength{\tabcolsep}{1mm}
\begin{tabular}{p{0.55\linewidth}ccc}
\toprule
    \textbf{Method} & \textbf{Venue}  & \textbf{mAP $\uparrow$} & \textbf{AUC $\uparrow$} \\ \midrule
 TSN-ResNet101~\cite{luo_arbee_2020} & IJCV &17.04 &62.29 \\ 
    I3D~\cite{luo_arbee_2020}  & IJCV & 15.37 &61.24 \\
    TSN~\cite{luo_arbee_2020}  & IJCV & 17.02 &62.70  \\
    Filntisis~\etal\cite{filntisis_emotion_2020}  & ECCV-W  &  16.56   &  62.66    \\
    Pikoulis~\etal~\cite{pikoulis_leveraging_2021} & FG &  19.29   &  66.82    \\ 
    EmotionCLIP~\cite{zhang2023learning} & ICCV  &     22.51 &   \underline{69.30} \\
    \textbf{Ours} ($E_V$ frozen) &  &  \underline{23.53}   &  67.62   \\
    \textbf{Ours}    &     &       \textbf{26.66}    & \textbf{69.83} \\
\bottomrule
\end{tabular}
\caption{BoLD~\cite{luo_arbee_2020}}
\label{tab:comparison_bold}
\end{subtable}
\begin{subtable}[t]{0.5\textwidth}\centering
\small
\setlength{\tabcolsep}{1mm}
\begin{tabular}{p{0.55\linewidth}cc}
\toprule
    \textbf{Method} & \textbf{Venue} & \textbf{Acc (\%) $\uparrow$} \\ \midrule
    CAER-Net~\cite{lee2019context}   & ICCV      & 73.47          \\
    EMOT-Net~\cite{kosti2017emotic}   & CVPR     & 74.51          \\
    GNN-CNN~\cite{zhang2019contextaware} & ICME  & 77.21          \\
    SIB-Net~\cite{li2021sequential}   & IJCB     & 74.56          \\
    RRLA~\cite{li2021human}     & IEEE T.Aff.C.          & 84.82          \\
    EmotiCon~\cite{mittal_emoticon_2020} & CVPR  & 88.65          \\
    EmotiCon+CCIM~\cite{yang_context_2023} & CVPR & \underline{91.17}\\
    EmotiCon+CLEF~\cite{yang2024robust} & CVPR    & 90.62          \\
    \textbf{Ours}            &              & \textbf{93.08} \\
\bottomrule
\end{tabular}
\caption{CAER-S~\cite{lee2019context}}
\label{tab:comparison_CAERS}
\end{subtable}
\begin{subtable}[t]{0.5\textwidth}\centering
\small
\setlength{\tabcolsep}{1mm}
\begin{tabular}{p{0.60\linewidth}cc}
\toprule
\textbf{Method}       & \textbf{Venue}         & \textbf{mAP $\uparrow$} \\ \midrule
EMOTNet~\cite{kosti2017emotic}       & CVPR        & 27.93 \\
GCN-CNN~\cite{zhang2019contextaware}    & ICME          & 28.16 \\
CAERNet~\cite{lee2019context}       & ICCV        & 23.85 \\
EmotiCon~\cite{mittal_emoticon_2020}   & CVPR           & 35.28 \\
RRLA~\cite{li2021human}        & IEEE T.Aff.C.          & 32.41 \\
VRD~\cite{hoang2021context}     & IEEE Access               & 35.16 \\
Costa~\etal~\cite {de_Lima_Costa_2023_CVPR}  & CVPR-W               & 30.02 \\
EmotiCon+CCIM~\cite{yang_context_2023}   & CVPR     & \textbf{39.13} \\
EmotiCon+CLEF~\cite{yang2024robust}  & CVPR      & 38.05 \\
\textbf{Ours}           &       & \underline{38.52} \\
\bottomrule
\end{tabular}
\caption{Emotic~\cite{kosti2017emotic}}
\label{tab:comparison_emotic}
\end{subtable}
\caption{Comparisons to the state-of-the-art across datasets.}
\label{tab:comparison_all}
\end{table}

In ~\cref{sec:sota}, we evaluate our method against state-of-the-art approaches across three emotion-in-context datasets, i.e CAER-S~\cite{lee2019context}, EMOTIC~\cite{kosti2017emotic}, and BoLD~\cite{luo_arbee_2020}. In ~\cref{ssec:ablation}, we perform an ablation study to assess the contribution of each component of our approach. In ~\cref{ssec:qualitative}, we provide an error analysis and discussion of our results.

\textbf{Datasets and Evaluations Metrics.} BoLD~\cite{luo_arbee_2020} is a video dataset of in-context human emotion recognition in the wild consisting of $9827$ video clips and $13239$ subjects each labeled with a bounding box and  $26$ discrete emotion categories. CAER-S consists of 70k static images annotated with the seven basic emotions: anger, disgust, fear, happiness, sadness, surprise, and neutral. EMOTIC is composed of $23571$ images with $34320$ annotated subjects in the wild. The annotations include bounding boxes and the annotation for 26 categorical emotions. For evaluation, we adhere to the metrics established by each respective dataset and report mean Average Precision (mAP) on EMOTIC and Bold. For the CAER-S, the standard classification accuracy is used for evaluation. 

\textbf{Implementation Details.} We initialise our backbone with the publicly available weights of InstructBLIP~\cite{dai2023instructblip}\footnote{We used the Salesforce/instructblip-flan-t5-xl checkpoint from huggingface. \url{https://huggingface.co/}} without the weights of the LLM. For training, we use the AdamW~\cite{loshchilov2018decoupled} optimiser.
Our framework is trained for a maximum of 50 epochs with early stopping and a linear rate scheduler.
For experiments on CAER-S and BoLD, we use weight decay of $0.1$. Experiments in EMOTIC use a weight decay of 0.0005 and a frozen vision encoder. As the weights of the backbone are initialised using a pre-trained network, we use variable learning rates for the classification layer and the backbone. Specifically, for CAER-S, we use an initial learning rate of $10^{-3}$, and for Emotic and BoLD, we use an initial learning rate of $10^{-4}$. The Q-Former and vision encoder learning rates are then adjusted by a factor of $0.1$. For BoLD, the vision encoder learning rate is lower and is adjusted by a factor of $0.01$. Finally, for experiments on BoLD, which is a video dataset, we sample 8 frames from each video and average pool their representations before passing them to the Q-Former.
The batch size is set to 64 for all static datasets and 4 for dynamic datasets.  As affective datasets are typically small and large models have a tendency to overfit, we add dropout in the attention layers with a probability of $0.4$ on the Q-Former and $0.3$ on the vision encoder.



\begin{table*}[ht!]
\centering
\small
\setlength{\tabcolsep}{1mm}
\begin{minipage}[t]{0.48\linewidth}
\begin{subtable}[t]{\linewidth}\centering
\begin{tabular}{p{0.35\linewidth}cccc}
    \toprule
    \textbf{Input} & \textbf{Train Vis.} & \textbf{Train Text} & \textbf{mAP$\uparrow$} & \textbf{AUC$\uparrow$} \\
    \hline
    \textbf{Ours} & \checkmark & \checkmark & \textbf{26.66} & \textbf{69.83} \\ 
    Ours & \xmark & \checkmark & 23.53 & 67.62 \\ \hline
    Ours(No bBox) & \xmark & \checkmark & 22.99 & 66.95 \\
    Ours(Subject) & \xmark & \checkmark & 23.75 & 66.26 \\
    Ours(No bBox) & \checkmark & \checkmark & 24.50 & 68.75 \\
    Ours(Subject) & \checkmark & \checkmark & 25.23 & 67.65 \\
    \bottomrule
\end{tabular}
\caption{Ablation study on the impact of training the vision encoder and the effect of bounding boxes. The top split evaluates the effect of training the vision encoder, while the bottom split further examines the influence of bounding boxes. "Train Vision" and "Train Text" indicate whether the vision and/or text modality is trained. "No bBox" refers to removing bounding box information, while "Subject" refers to using the cropped subject instead of full-image context.}
\label{tab:ours_bold}
\end{subtable}
\end{minipage}
\hfill
\begin{minipage}[t]{0.48\linewidth}
\begin{subtable}[t]{\linewidth}\centering
\begin{tabular}{p{0.40\linewidth}ccc}
    \toprule
    \textbf{Architecture} & \textbf{Input type} & \textbf{mAP$\uparrow$} & \textbf{AUC$\uparrow$} \\
    \hline
    $E_V$(whole image) & Vis. & 16.26 & 63.38 \\
    $E_V^*$(whole image) & Vis. & 16.42 & 60.89 \\
    $E_V$(subject) & Vis. & 16.18 & 61.49 \\ \hline
    LlaVa-1.5, zero-shot\dag & Text & 11.43 & 50.22 \\
    LlaVa-1.5, LP + LoRA & Text, Vis. & 26.25 & 68.74\\
    RoBERTa\ddag & Text & 17.45 & 60.24 \\ \hline
    $Cat(E_V, RoBERTa)\ddag$ & Text, Vis. & 17.74 & 61.75 \\ 
    $Cat(E_V^*, RoBERTa)\ddag$ & Text, Vis. & 17.21 & 60.23 \\ 
    \bottomrule
\end{tabular}
\caption{Ablation on the impact of the text input, visual input, and their fusion. ``Text'' and ``Vis'' indicate whether text and/or vision inputs are used. \\ $*$ denotes a frozen model. \\ \dag Zero-shot evaluation of the generation method \\ \ddag As our architecture lacks a text modality, we use RoBERTa as a text backbone in the ablations.}
\label{tab:modalities_bold}
\end{subtable}
\end{minipage}

\begin{minipage}[t]{0.48\linewidth}
\begin{subtable}[t]{\textwidth}\centering
\begin{tabular}{p{0.55\linewidth}cc}
    \toprule
   \textbf{ Method } & \textbf{mAP $\uparrow$}  & \textbf{AUC $\uparrow$} \\ \midrule
    Ours (Transformer-based) & 23.9 & 67.0 \\
    \textbf{Ours} (Temporal Avg Pooling) & \textbf{26.66} & \textbf{69.83} \\
    \bottomrule
\end{tabular}
\caption{Ablation on the temporal modeling strategy. ``Transformer-based '' refers to modeling the temporal dimension using a Transformer encoder module. ``Temporal Avg Pooling'' refers to token averaging tokens over time.}
\label{tabl:boldtemp}
\label{tab:comparison_bold_temp}
\end{subtable}
\end{minipage}
\hfill
\begin{minipage}[t]{0.48\linewidth}
\begin{subtable}[t]{\textwidth}\centering
\begin{tabular}{p{0.55\linewidth}cc}
    \toprule
   \textbf{ \#Frames ($T$)} & \textbf{mAP $\uparrow$}  &\textbf{ AUC $\uparrow$} \\ \midrule
    4 & 26.51 & \textbf{69.85} \\
    \textbf{8} & \textbf{26.66} & 69.83 \\
    16 & \underline{26.60} & \underline{69.80} \\
    \bottomrule
\end{tabular}
\caption{Ablation on the effect of the number of frames $T$ per clip.}
\label{tab:boldframes}
\end{subtable}
\end{minipage}

\caption{Ablation Study of Model Components on the Multi-Label Emotion Classification of the BoLD~\cite{luo_arbee_2020} Dataset}
\label{tab:boldablation}
\end{table*}

\subsection{Comparison with Other State-Of-The-Art}
\label{sec:sota}

In ~\cref{tab:comparison_all}, we demonstrate our model's performance against other SoTA. For BoLD, our method significantly outperforms the state-of-the-art, achieving over $4\%$ higher mAP than the previous best. In CAER-S, our method surpasses the previous SoTA by nearly $2\%$ in terms of accuracy. In EMOTIC, our method performs comparably to the previous SOTA.

\subsection{Ablation Studies}
\label{ssec:ablation}

For ablations, we denote as \textbf{Ours}, our full model that integrates all components described in~\cref{sec:architecture}, achieving an mAP of $26.66$ and an AUC $69.83$. We analyze the influence of individual model components by examining the impact of bounding boxes (~\cref{tab:ours_bold}). We then investigate the effect of text and visual features (~\cref{tab:modalities_bold}). Finally, we explore the effect of frame sampling and temporal modeling (~\cref{tabl:boldtemp,tab:boldframes}).

\textbf{Effect of Training Vis. Encoder.} We evaluate the impact of training the vision encoder, as shown in in~\cref{tab:ours_bold} (top split). Our results show that fine-tuning the vision encoder improves performance compared to keeping it frozen, increasing mAP by 3.13\% and AUC by 2.21\%. This highlights the importance of refining visual representations alongside text-based reasoning in vision-language models for this task.

\textbf{Bounding Boxes.} 
As described in the method section, to differentiate individuals in visual input, we a) prompt the VLLM specifically about the person within the red bounding box and b) overlay a bounding box around the subject in each sample so that it is used as RGB input. To assess the influence of localising a person using a bounding box, we conduct two experiments: a) using images without bounding boxes and b) cropping the subject of interest. Our findings, shown in~\cref{tab:ours_bold}, reveal that using the red bounding box improves our pipeline, increasing mAP by $2.16\%$ and AUC by $1.08\%$. However, cropping the image significantly decreases performance, underscoring the importance of aligning text descriptions with visual input.

\textbf{Visual Features.} 
We evaluate the vision modality's effectiveness using a standard linear probe approach by adding a single classification layer over our vision encoder, $E_v$. We examine three cases: training the entire pipeline end-to-end (first-row in~\cref{tab:modalities_bold}), freezing the backbone and only training the classifier (second row), and training end-to-end with the image cropped to show only the subject of interest (last row). Our results show that visual features alone are insufficient for good performance, likely due to noise and bias in the image, as indicated by a more than $10\%$ lower mAP without the use of generated text context.

\begin{table}[t]
\centering
\begin{subtable}[t]{\linewidth}\centering
\small
    \begin{tabular}{p{0.4\linewidth}cl}
        \toprule
        \textbf{Architecture}          & \textbf{Input type} & \textbf{mAP$\uparrow$}   \\
        \midrule
        $E_V$                          & Vis.                & 91.60       \\
        $E_V^*$                        & Vis.                & 49.43                \\ 
        \midrule
        LlaVa-1.5, zero-shot\dag       & Text                & 15.29              \\
        LlaVa-1.5, LP + LoRA & Text, Vis. & 92.95\\
        RoBERTa\ddag                   & Text                & 45.36              \\ 
        \midrule
        $Cat(E_V^*, RoBERTa)\ddag$     & Text, Vis.          & 53.78            \\ 
        $Cat(E_V, RoBERTa)\ddag$       & Text, Vis.          & 92.07               \\ 
        \midrule
        \textbf{Ours}                   & Text, Vis.         & \textbf{93.08}      \\ 
        \bottomrule
    \end{tabular}
\caption{CAER-S}
\label{tab:modalities}
\end{subtable}

\begin{subtable}[t]{\linewidth}\centering
\small
    \begin{tabular}{p{0.4\linewidth}cl}
        \toprule
        \textbf{Architecture}             & \textbf{Input type}   & \textbf{mAP$\uparrow$}   \\
        \midrule
        $E_V$ (whole image)               & Vis.                  & 29.19      \\
        $E_V^*$ (whole image)             & Vis.                  & 27.17                \\
        $E_V$ (subject)                   & Vis.                  & 25.55               \\ 
        \midrule
        LlaVa-1.5, zero-shot\dag    & Text & 16.98            \\
        LlaVa-1.5, LP + LoRA & Text, Vis. & 38.11\\
        RoBERTa\ddag                    & Text & 34.25           \\    
        \midrule
        $Cat(E_V, RoBERTa)\ddag$          & Text, Vis.            & 34.66             \\ 
        $Cat(E_V^*, RoBERTa)\ddag$        & Text, Vis.            & 34.29                  \\ 
        \midrule
        \textbf{Ours}                   & Text, Vis.         & \textbf{38.52}      \\ 
        \bottomrule
    \end{tabular}
\caption{EMOTIC}
\end{subtable}
\begin{subtable}[t]{\linewidth}\centering
\small 
    \begin{tabular}{p{0.3\linewidth}ccc}
        \toprule
        \textbf{Input}            & \textbf{T. Vision} & \textbf{T. Text}   & \textbf{Acc$\uparrow$}   \\
        \midrule
        \textbf{Ours}       & \checkmark         & \checkmark         & \textbf{93.08}          \\
        Ours              & \xmark             & \checkmark         & 88.76                   \\  
        \bottomrule
    \end{tabular}
\caption{CAER-S}
\label{tab:ours}
\end{subtable}

\begin{subtable}[t]{\linewidth}\centering
\small 
    \begin{tabular}{p{0.45\linewidth}ccc}
        \toprule
        \textbf{Input}                     & \textbf{T. Vision} & \textbf{T. Text}          & \textbf{mAP$\uparrow$}   \\
        \midrule
        \textbf{Ours}    & \xmark             & \checkmark               & \textbf{38.52}          \\ 
        Ours (\small{GPT4V Desc.})       & \xmark             & \checkmark               & 37.90                   \\ 
        Ours            & \checkmark         & \checkmark               & 38.09                   \\ 
        \midrule
        Ours (\small{No bounding Box})             & \xmark             & \checkmark               & 37.61                   \\
        Ours (\small{Subject})                     & \xmark             & \checkmark               & 37.88                   \\
        \bottomrule
    \end{tabular}
\caption{EMOTIC}
\label{tab:oursemotic}
\end{subtable}
\caption{Ablation Study of Model Components on the EMOTIC~\cite{kosti2017emotic} and CAER-S~\cite{lee2019context}}
\label{tab:whole_table}
\end{table}

\textbf{Text Features.}
We test if our description extraction model, LlaVa-1.5~\cite{liu2023improvedllava}, can predict emotion categories in a zero-shot manner. We prompt LlaVa to predict the subject's apparent emotion in a multiple-choice setting using contextual cues. Accuracy is measured by checking if the response contains the class name~\cite{xu2023lvlm}. Our method is significantly outperforming LlaVa's zero-shot mAP by +$\sim15\%$ mAP as seen in~\cref{tab:modalities_bold}, highlighting the need for a task-specific model. This is likely due to challenges in multi-label tasks and noisy label definitions in emotion prediction. To see if text descriptions alone suffice for in-context emotion prediction, we use RoBERTa~\cite{liu2019roberta} as a language baseline, as our architecture does not have a text modality (the Q-Former takes text embeddings as input). RoBERTa's last hidden state of the class token is fed to a classifier; as shown in~\cref{tab:modalities_bold}, text features alone perform worse than our fusion pipeline. 

\textbf{Fusion.}
To investigate the Q-Former fusion method, we compare it with a baseline model that concatenates text features from RoBERTa and visual features from our vision encoder $E_v$ before classification. As shown in~\cref{tab:modalities_bold}, our Q-Former approach is more effective (+$\sim10\%$ mAP and +$\sim9\%$ AUC). Additionally, we trained the pipeline with the visual modality frozen and found that fine-tuning the visual modality has minimal impact. Finally, we add a Linear Probe (LP) layer and fine-tune the LlaVa architecture using LoRA~\cite{hu2022lora} adapters, using the images and instructing the model to predict emotion. 
While the extensive pre-training of LlaVa endows the model with strong commonsense reasoning capabilities, its large size complicates fine-tuning and challenges our goal of maintaining a streamlined approach.

\begin{figure*}[ht]
\centering
  \begin{subfigure}[t]{0.24\textwidth}  
      \subcaption[labels]{\textbf{Labels}: `Affection', `Happiness', `Pleasure'}
      \centering
        \includegraphics[width=0.6\textwidth]{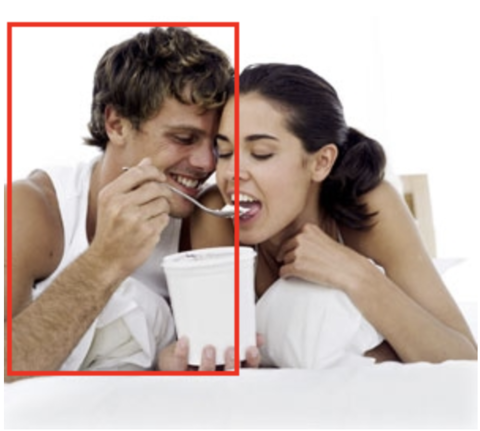}
        
      {\tiny \textit{The most suitable emotions to describe how the person inside the red box feels are "Affection" and "Happiness." The man and woman are cuddling and sharing a spoonful of ice cream, which suggests a close and loving relationship between them. Their smiles and the act of eating ice cream together indicate a sense of happiness and enjoyment in each other's company.}
      \\Ours:  \colorbox{lime}{`Affection'}, \colorbox{lime}{`Happiness'}, \colorbox{lime}{`Pleasure'}
      \\$E_V(cropped)$: \colorbox{pink}{[]}
      \\$E_V(whole)$: \colorbox{pink}{`Anger'}, \colorbox{pink}{`Annoyance'}, \colorbox{pink}{`Aversion'}, \colorbox{pink}{`Disapproval'}, \colorbox{pink}{`Disconnection'}, \colorbox{pink}{`Disquietment'}, \colorbox{pink}{`Doubt/Confusion'}, \colorbox{pink}{`Engagement'}, \colorbox{pink}{`Fatigue'}, \colorbox{pink}{`Happiness'}, \colorbox{pink}{`Pain'}, \colorbox{pink}{`Sadness'}, \colorbox{pink}{`Suffering'}, \colorbox{pink}{`Surprise'}}
  \end{subfigure}
  \begin{subfigure}[t]{0.24\textwidth}  
      \subcaption[labels]{\textbf{Labels}: `Anticipation', `Confidence', `Engagement', `Sympathy'}
      \centering
      \includegraphics[width=0.6\textwidth]{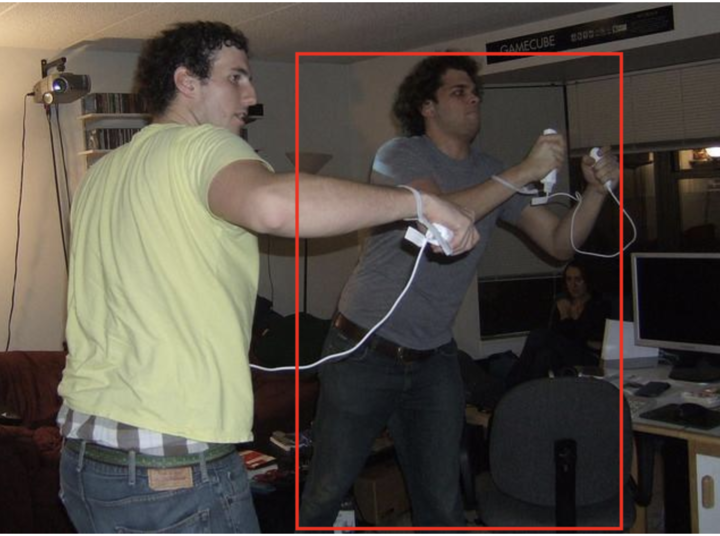}
      
      {\tiny 
        \textit{The person inside the red box appears to be experiencing a mix of emotions. They are holding a Wii remote, which suggests that they are engaged in playing a video game. This could evoke feelings of excitement, anticipation, and engagement. However, the person is also making a funny face, which indicates that they might be experiencing humor, amusement, or playfulness. The combination of these emotions suggests that the person inside the red box is having a fun and entertaining time while playing the video game.}
        \\Ours: \colorbox{lime}{`Anticipation'}, \colorbox{lime}{`Confidence'}, \colorbox{lime}{`Engagement'}, \colorbox{lime}{`Excitement'}
        \\$E_V(cropped)$: \colorbox{lime}{`Engagement'}
        \\$E_V(whole)$: \colorbox{lime}{`Anticipation'}, \colorbox{lime}{`Confidence'}, \colorbox{pink}{`Disconnection'}, \colorbox{lime}{`Engagement'}, \colorbox{pink}{`Esteem'}, \colorbox{pink}{`Excitement'}, \colorbox{pink}{`Happiness'}, \colorbox{pink}{`Peace'}, \colorbox{pink}{`Pleasure'}, \colorbox{lime}{`Sympathy'}
       }
  \end{subfigure}
  \begin{subfigure}[t]{0.24\textwidth}  
      \subcaption[labels]{\textbf{Labels}: `Anticipation', `Confidence', `Engagement', `Excitement'}
      \centering
      \includegraphics[width=0.6\textwidth]{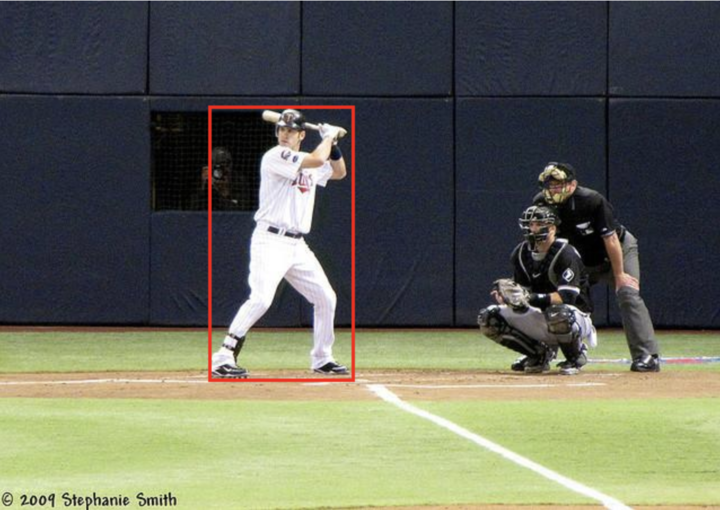}
      
      {\tiny 
        \textit{ The most suitable emotions to describe how the person inside the red box feels are "Confidence" and "Excitement." The man is holding a baseball bat and appears to be preparing to swing, which suggests that he is confident in his ability to hit the ball and excited about the game. Other emotions, such as "Anger" or "Disapproval," might not be as fitting, as the image does not show any signs of aggression or disapproval.}
          \\Ours: \colorbox{lime}{`Confidence'}, \colorbox{lime}{`Engagement'}, \colorbox{lime}{`Excitement'}
          \\$E_V(cropped)$: \colorbox{lime}{`Engagement'}
          \\$E_V(whole)$: \colorbox{lime}{`Anticipation'}, \colorbox{lime}{`Confidence'}, \colorbox{pink}{`Disquietment'}, \colorbox{lime}{`Engagement'}, \colorbox{pink}{`Esteem'}, \colorbox{lime}{`Excitement'}, \colorbox{pink}{`Happiness'}, \colorbox{pink}{`Pleasure'}
      }
  \end{subfigure}
  \begin{subfigure}[t]{0.24\textwidth}  
      \subcaption[labels]{\textbf{Labels}: `Confidence', `Engagement', `Esteem', `Excitement'}
      \centering
      \includegraphics[width=0.6\textwidth]{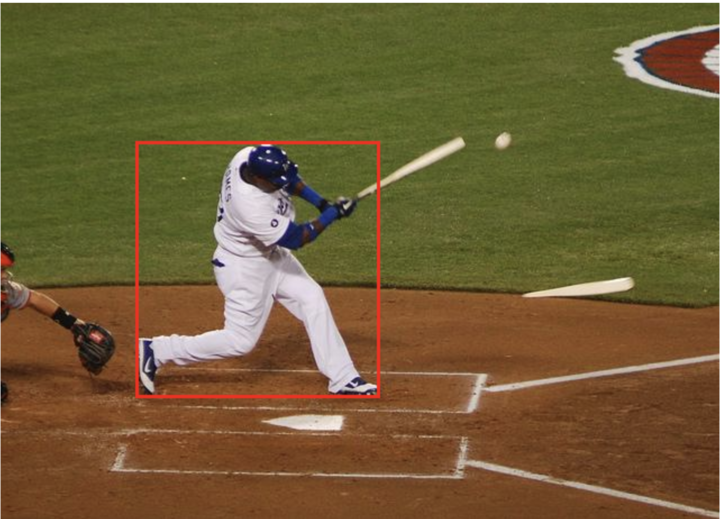}
      
      {\tiny \textit{The person inside the red box is a baseball player swinging a bat to hit a ball. Given the list of emotions provided, the most suitable emotions to describe how the person feels are "Confidence" and "Excitement." The player's posture and focus on the ball suggest that they are confident in their ability to hit the ball and excited about the potential outcome of their swing. Other emotions, such as "Anger" or "Fear," may not be as applicable in this context, as the player appears to be calm and focused on the task at hand.}
      \\Ours: \colorbox{lime}{`Confidence'}, \colorbox{lime}{`Engagement'}, \colorbox{lime}{`Excitement'}
      \\$E_V(cropped)$: \colorbox{pink}{`Anticipation'}, \colorbox{lime}{`Engagement'}
      \\$E_V(whole)$: \colorbox{pink}{`Anticipation'}, \colorbox{lime}{`Confidence'},  \colorbox{lime}{`Engagement'}, \colorbox{pink}{`Esteem'}, \colorbox{lime}{`Excitement'}, \colorbox{pink}{`Happiness'}, \colorbox{pink}{`Pleasure'}}
  \end{subfigure}

    \caption{Qualitative comparison of predictions made by our proposed method vs a vision encoder trained on the cropped image of the subject ($E_V(cropped)$) and whole images ($E_V(whole)$) --including the bounding box and visual context, from the test set of the EMOTIC~\cite{kosti2017emotic} dataset.}
  \label{fig:qualitative}
\end{figure*}

\textbf{Temporal Dimension.} As the BoLD~\cite{luo_arbee_2020} dataset contains video clips, we compare our method of averaging visual tokens with an architecture that uses two Transformer Encoder layers as a temporal module. Since all visual tokens are processed, the temporal module must handle multiple tokens per frame. We achieve this by replicating the temporal module for each $\frac{H}{P} \times \frac{W}{P} + 1$ tokens, sharing weights. The results are shown in~\cref{tabl:boldtemp}. Notably, we observe that simple average pooling is sufficient to model in-context emotion over time. Additionally, in~\cref{tab:boldframes}, we ablate the number of frames used per clip.

\textbf{Interpretation of Results.}  We note that incorporating generated contextual descriptions achieves +10\% mAP on two highly contextual datasets, BoLD and EMOTIC, compared to using visual information alone. 
For the less contextually intensive CAER-S dataset, the addition of generated contextual descriptions provides a smaller mAP improvement of +2\%. This showcases that the generated contextual descriptions can effectively enhance context modelling.  

\textbf{Additional ablations.} In~\cref{tab:whole_table}, we present additional ablation studies evaluating the impact of input modalities and fusion methods across the EMOTIC and CAER datasets. Single modalities generally exhibit sub-optimal performance.  However, the language-guided model achieves significantly better results by combining these features, which corroborates findings in our main ablation. 

For the BoLD and EMOTIC datasets, our approach yields approximately 24\% and 39\% improvements, respectively, while for CAER-S, the increase is around 2\%. We attribute the substantial gains in BoLD and EMOTIC to the highly contextual nature of these datasets. In~\cref{tab:ours} and ~\cref{tab:oursemotic}, we present ablation studies on the quality of text descriptions and the impact of bounding boxes on CAER-S and EMOTIC datasets.

Additionally, for the EMOTIC dataset, we use GPT4v-generated descriptions.  The results of our architecture trained with GPT4v descriptions can be seen in~\cref{tab:oursemotic}.
 Our architecture yields an mAP of $37.9\%$, $0.5$ lower than but comparable to LlaVa descriptions, highlighting that our method -- using VLLM-generated context-- is robust to input. Interestingly, the GPT4V responses are shorter and less descriptive, which has been previously observed in other tasks\footnote{\url{https://llava-vl.github.io/}}. We theorise that the shorter descriptions include less of the image context and are the reason for the lower performance.

\subsection{Qualitative analysis and discussion}
\label{ssec:qualitative}
\begin{figure}[h]
\centering
\begin{subfigure}[b]{0.45\textwidth}
   \includegraphics[width=1\linewidth]{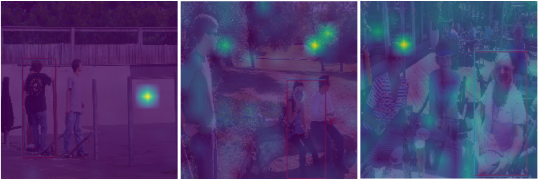}
   \caption{Vision Encoder trained on the  whole images}
   \label{fig:attentionblip} 
\end{subfigure}

\begin{subfigure}[b]{0.45\textwidth}
   \includegraphics[width=1\linewidth]{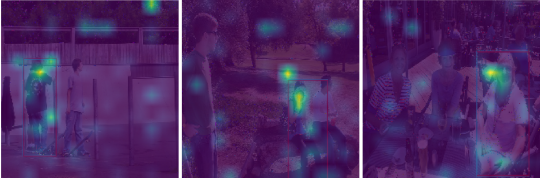}
   \caption{Proposed method}
   \label{fig:attentionours}
\end{subfigure}
    \caption{Examples of Attention Maps of the last attention layer of the Vision Encoder only trained on the whole image --\ie subject and context (top), and the last Q-Former cross-attention in our proposed method (bottom).}
  \label{fig:attention}
\end{figure}

\subsubsection{Model's Predictions}

To qualitatively assess our model's predictions, we compare its output with that of the ViT vision encoder ($E_V$) trained under two conditions: without context on subject images ($E_V(cropped)$) and with the entire image including visual context and bounding box ($E_V(whole)$).~\cref{fig:qualitative} shows examples from the EMOTIC~\cite{kosti2017emotic} test set for easier visualisation, along with predictions from our model and the two ViT architectures.
$E_V(cropped)$ had many false negatives, often failing to predict any classes confidently. $E_V(whole)$ predicted many classes, causing numerous false positives. The rich but noisy visual context led to many predictions without meaningful patterns. Consistency analysis showed that our method and the $E_V(whole)$ had more consistent predictions than $E_V(cropped)$ when shown similar images (~\cref{fig:qualitative}).

Finally, we compare the attention maps from the $E_V(whole)$ and our proposed method. Specifically, we analyse the last self-attention layer of the vision-only model and the last cross-attention layer of our method. As shown in~\cref{fig:attention}, our method's cross-attention mechanism focuses more on the subject, while the $E_V$ attention is more dispersed, despite our model not being trained on localisation.

\subsubsection{Error Analysis: The Role of Context and Ambiguity in Emotion Recognition}



\begin{table}[]
\centering
\large
\resizebox{\linewidth}{!}{%
\begin{tabular}{lccc}
\toprule
 \begin{tabular}[c]{@{}c@{}}\ \textbf{IoU} \\ \textbf{Thresholds }\end{tabular} & \begin{tabular}[c]{@{}c@{}}\ \textbf{mAP@xIoU} \\\end{tabular}&  \begin{tabular}[c]{@{}c@{}}\ \textbf{mAP} \\\textbf{Remaining} \end{tabular} & \begin{tabular}[c]{@{}c@{}} \textbf{\# Samples} \\ \textbf{(Overlapping / Remaining)}\end{tabular} \\ \hline
$0.2$ & $41.34$       & $36.81$     & $1104$ / $6099$                     \\
$0.3$ & $45.32$       & $36.65$     & $511$ / $6692$                      \\
$0.4$ & $45.03$       & $37.11$     & $269$ / $6934$                      \\
$0.5$ & $44.02$       & $37.68$     & $151$ / $7052$                      \\
$0.7$ & $44.42$       & $37.93$     & $61$ / $7142$                       \\
\bottomrule
\end{tabular}%
}
\caption{Mean Average Precision (mAP) of bounding boxes categorized as "Overlapping" (IoU with another box above the threshold) vs. "Remaining" (not overlapping), across different IoU thresholds. The last column shows the number of samples in each category.}
\label{tab:iou}
\end{table}

\begin{figure*}[h]
\pgfdeclarelayer{background layer}
\pgfsetlayers{background layer, main}
    \begin{tikzpicture}[node distance=.1cm]
        \tiny
        \node[text width=.3\textwidth, align=left] (image_A) {\includegraphics[width=\linewidth]{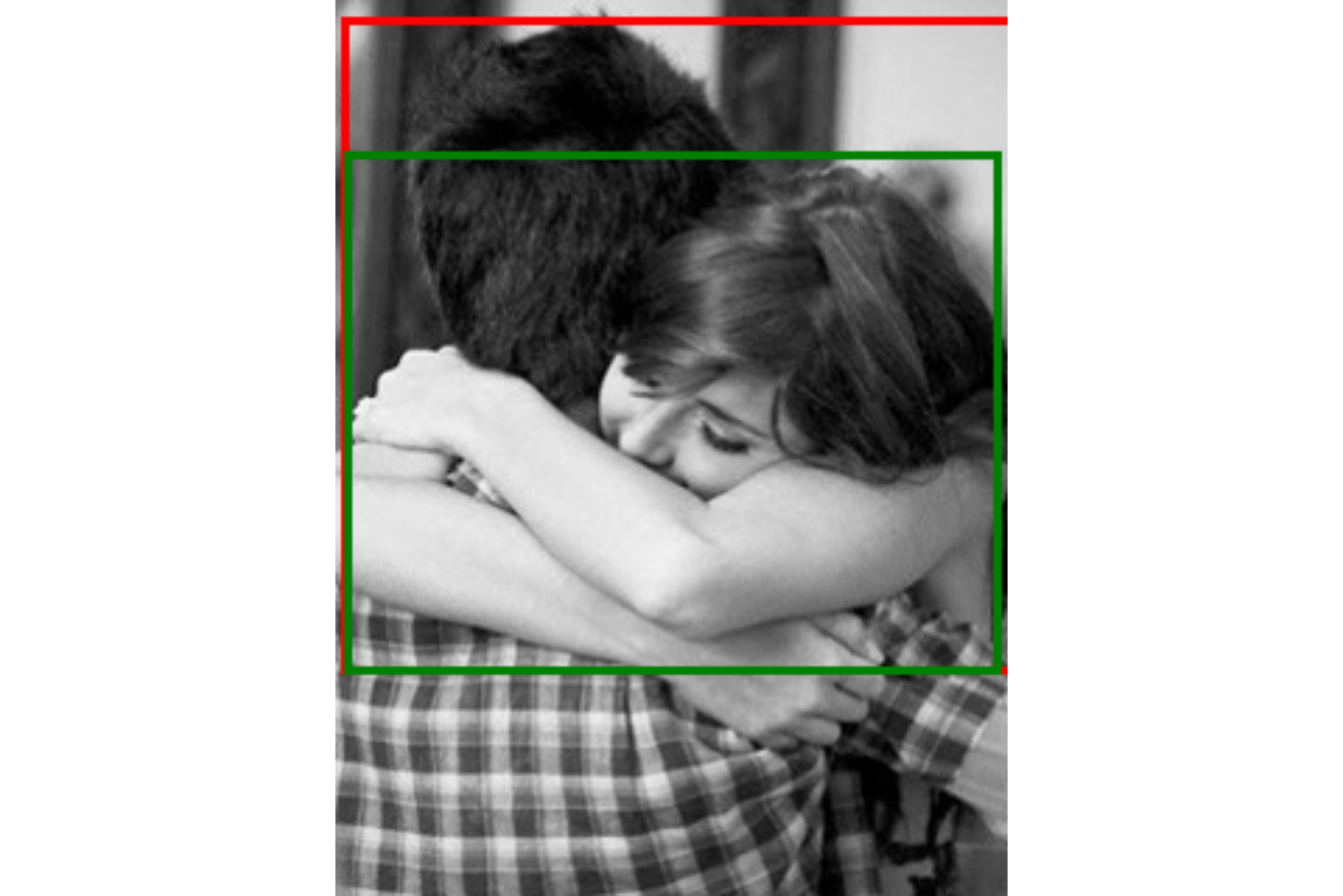}};
        \node[text width=.3\textwidth, align=left, right=of image_A] (image_B) {\includegraphics[width=\linewidth]{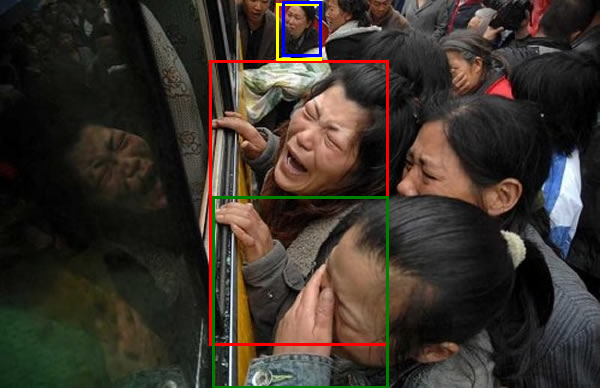}};
        \node[text width=.3\textwidth, align=left, right=of image_B] (image_C) {\includegraphics[width=\linewidth]{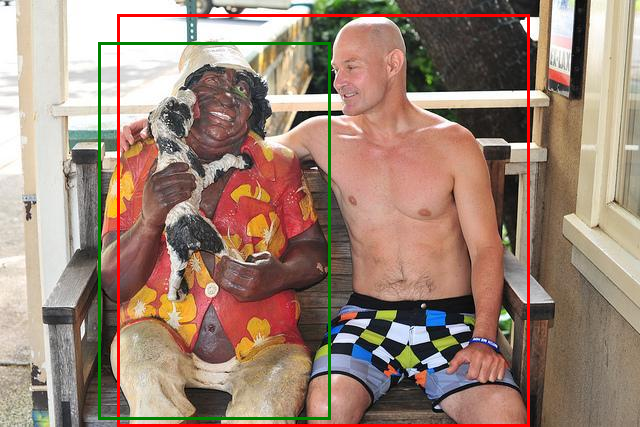}};
        \node[text width=.3\textwidth, align=left, below=of image_A] (label_a1) {
            \textbf{\color{rose}{Red Box} Ground Truth:} `Confidence', `Engagement', `Happiness', `Sympathy'
        };
        \node[text width=.3\textwidth, align=left, below=of label_a1] (pred_a1) {
            \textbf{Prediction:} \colorbox{pink}{`Affection'}, \colorbox{lime}{`Happiness'}, \colorbox{pink}{`Pleasure'}
        };
        \node[text width=.3\textwidth, align=left, below=of pred_a1] (desc_a1) {
            \textbf{Pseudo-caption:} Based on the image context, the most suitable emotions to describe how the person inside the red box feels are ``Affection'', ``Happiness'', ``Peace'', and ``Sadness''. The man and woman are embracing each other, which suggests a strong emotional connection and bond between them. Their body language and facial expressions indicate that they are experiencing positive emotions, such as affection, happiness, and peace, as well as a sense of sadness or longing. It is important to note that emotions can be complex and multifaceted, so while these are the most suitable emotions based on the image context, it is possible that other emotions may also be present.
        };
        \begin{pgfonlayer}{background layer}
                \draw [draw=none, rounded corners=1mm, baseline, fill=rose!15] (label_a1.north west) rectangle (desc_a1.south east);
        \end{pgfonlayer}
        \node[text width=.3\textwidth, align=left, below=of desc_a1] (label_a2) {
            \textbf{\color{forest}{Green Box Ground Truth:} }`Affection', `Excitement', `Happiness', `Peace'
        };
        \node[text width=.3\textwidth, align=left, below=of label_a2] (pred_a2) {
            \textbf{Prediction:} \colorbox{lime}{`Affection'}, \colorbox{lime}{`Happiness'}, \colorbox{pink}{`Pleasure'}, \colorbox{lime}{`Sympathy'}
        };
        \node[text width=.3\textwidth, align=left, below=of pred_a2] (desc_a2) {
            \textbf{Pseudo-caption:} Based on the image context, the most suitable emotions to describe how the person inside the red box feels are ``Affection'', ``Happiness'', ``Peace'', and ``Sadness''. The man and woman are hugging each other, which indicates a strong emotional connection and affection between them. Their body language and facial expressions suggest that they are experiencing happiness, peace, and a sense of comfort in each other's presence. {\color{deepsea} Additionally, the woman appears to be crying, which implies that she might be experiencing sadness or a mix of emotions.}
        };
        \begin{pgfonlayer}{background layer}
                \draw [draw=none, rounded corners=1mm, baseline, fill=forest!15] (label_a2.north west) rectangle (desc_a2.south east);
        \end{pgfonlayer}
        \node[text width=.3\textwidth, align=center, below=of desc_a2] (A) {\centering\normalsize{(a)}};
        
        \node[text width=.3\textwidth, align=left, below=of image_B] (label_b1) {
            \textbf{\color{rose}{Red Box Ground Truth:}} `Disapproval', `Pain', `Sadness', `Suffering'
        };
        \node[text width=.3\textwidth, align=left, below=of label_b1] (pred_b1) {
            \textbf{Prediction:} \colorbox{lime}{`Pain'}, \colorbox{lime}{`Sadness'}, \colorbox{lime}{`Suffering}
        };
        \node[text width=.3\textwidth, align=left, below=of pred_b1] (desc_b1) {
            \textbf{Pseudo-caption:} Based on the image context, the person inside the red box appears to be experiencing a mix of emotions, including sadness, fear, and discomfort. The woman's facial expression and body language suggest that she is in distress, possibly due to being trapped in a confined space or experiencing an unpleasant situation. The presence of {\color{deepsea} a crowd of people surrounding the red box} may also contribute to her feelings of fear and discomfort. Overall, the emotions of sadness, fear, and discomfort are more suitable for describing how the person inside the red box feels in this context.
        };
        \begin{pgfonlayer}{background layer}
                \draw [draw=none, rounded corners=1mm, baseline, fill=rose!15] (label_b1.north west) rectangle (desc_b1.south east);
        \end{pgfonlayer}
        \node[text width=.3\textwidth, align=left, below=of desc_b1] (label_b2) {
            \textbf{\color{deepsea}{Blue Box Ground Truth:}} `Disquietment', `Sadness', `Sensitivity', `Suffering', `Sympathy'
        };
        \node[text width=.3\textwidth, align=left, below=of label_b2] (pred_b2) {
            \textbf{Prediction:} \colorbox{pink}{`Pain'}, \colorbox{lime}{`Sadness'}, \colorbox{lime}{`Suffering'}
        };
        \node[text width=.3\textwidth, align=left, below=of pred_b2] (desc_b2) {
            \textbf{Pseudo-caption:} Based on the image context, the person inside the blue box appears to be experiencing a mix of emotions such as sadness, fear, and discomfort. The woman's facial expression and body language suggest that she is feeling distressed, possibly due to the situation she is in or the presence of the crowd surrounding her. The presence of the crowd and {\color{deepsea} the fact that she is inside a blue box might also contribute to her feelings of fear and discomfort.} However, it is important to note that emotions can be complex and multifaceted, and the exact emotions experienced by the person inside the red box cannot be definitively determined from the image alone.
        };
        \begin{pgfonlayer}{background layer}
                \draw [draw=none, rounded corners=1mm, baseline, fill=deepsea!15] (label_b2.north west) rectangle (desc_b2.south east);
        \end{pgfonlayer}
        \node[text width=.3\textwidth, align=center, below=of desc_b2] (B) {\normalsize{(b)}};

        \node[text width=.3\textwidth, align=left, below=of image_C] (label_c1) {
            \textbf{\color{rose}{Red Box Ground Truth:}} `Pleasure', `Sympathy'
        };
        \node[text width=.3\textwidth, align=left, below=of label_c1] (pred_c1) {
            \textbf{Prediction:} \colorbox{pink}{`Happiness'}
        };
        \node[text width=.3\textwidth, align=left, below=of pred_c1] (desc_c1) {
            \textbf{Pseudo-caption:} Based on the image context, the person inside the red box appears to be experiencing a mix of positive and negative emotions. They are smiling, which suggests that they are experiencing happiness or contentment. However, {\color{deepsea} they are also holding a stuffed animal}, which could indicate that they are feeling a sense of nostalgia, comfort, or playfulness. {\color{deepsea} Additionally, they are sitting on a bench with a shirtless man,} which might suggest that they are feeling a sense of relaxation or enjoyment in each other's company. Overall, the person inside the red box seems to be experiencing a combination of happiness, contentment, nostalgia, comfort, and relaxation.
        };
        \begin{pgfonlayer}{background layer}
                \draw [draw=none, rounded corners=1mm, baseline, fill=rose!15] (label_c1.north west) rectangle (desc_c1.south east);
        \end{pgfonlayer}
        \node[text width=.3\textwidth, align=left, below=of desc_c1] (label_c2) {
            \textbf{\color{forest}{Green Box Ground Truth:}} `Affection', `Happiness', `Pleasure'
        };
        \node[text width=.3\textwidth, align=left, below=of label_c2] (pred_c2) {
            \textbf{Prediction:} \colorbox{lime}{`Happiness'}
        };
        \node[text width=.3\textwidth, align=left, below=of pred_c2] (desc_c2) {
            \textbf{Pseudo-caption:} Based on the image context, the person inside the green box appears to be feeling a mix of emotions. They are smiling, which suggests that they are experiencing happiness or pleasure. However, they are also holding a stuffed animal, which could indicate that they are feeling a sense of nostalgia, comfort, or affection towards the toy. Additionally, they are sitting on a bench, which might suggest that they are taking a break, resting, or simply enjoying their time outdoors. Overall, the person inside the red box seems to be experiencing a combination of happiness, comfort, and a sense of nostalgia or affection towards the stuffed animal.
        };
        \begin{pgfonlayer}{background layer}
                \draw [draw=none, rounded corners=1mm, baseline, fill=forest!15] (label_c2.north west) rectangle (desc_c2.south east);
        \end{pgfonlayer}
        \node[text width=.3\textwidth, align=center, below=of desc_c2] (C) {\normalsize{(c)}};
    \end{tikzpicture}
    \caption{Examples of samples with multiple subjects and overlapping bounding boxes.}
    \label{fig:error_iou}
\end{figure*}

As shown in \cref{fig:error_iou}, in-context emotion datasets often contain noise, including multiple subjects in close proximity, occluded faces, and overlapping bounding boxes, which introduce challenges for automated emotion recognition models. To better understand the model's behaviour in these ambiguous scenarios, we evaluate its performance on samples with overlapping bounding boxes by calculating mAP at various Intersection over Union (IoU) thresholds. \cref{tab:iou} presents the results across different mAP@xIoU thresholds. The second column reports the model's performance on samples with highly overlapping bounding boxes (IoU above a given threshold), while the third column corresponds to the performance on less overlapping or isolated samples. The final column indicates the proportion of samples detected at each IoU threshold.


A key observation from these results is that the model performs better when bounding boxes overlap. This suggests that shared emotional context among individuals in the same scene aids recognition—people in similar environments often express similar emotions. However, this reliance on context can also introduce ambiguity, making it difficult to generate distinct predictions for each individual.

To further analyze these errors, \Cref{fig:error_iou} examines model-generated descriptions for overlapping bounding boxes. Across all cases, descriptions for subjects in the same image are highly similar, reflecting the challenge of disambiguating individuals when bounding boxes overlap.

\begin{itemize}
    \item \textbf{Contextual Similarity:} In the first case, both subjects receive nearly identical descriptions and emotion predictions, reinforcing the model’s tendency to generalize emotions based on shared context.
    \item \textbf{Bounding Box Hallucination:} In the second case, the model mistakenly describes the bounding box itself as a physical object, though this error does not significantly impact performance.
    \item \textbf{Failure to Disambiguate:} In the third case, a bounding box enclosing multiple subjects prevents the model from distinguishing individual emotions, leading to ambiguous predictions.
\end{itemize}

These findings highlight the trade-off between leveraging contextual cues and preserving subject-specific differentiation, a key challenge in vision-language models for affective computing.

\section{Limitations and Considerations}

We propose leveraging VLLMs to enhance in-context emotion prediction by generating emotionally aware descriptions that support the classification process. While our method achieves state-of-the-art performance in this task, it has certain limitations. First, as demonstrated in our experiments, the quality of the generated textual descriptions is crucial for accurate predictions. Since these descriptions are directly dependent on the VLLM’s capabilities, any inaccuracies or ambiguities—such can negatively impact emotion classification. Additionally, the use of VLLMs in this context raises ethical concerns related to bias. LLMs are trained on vast, largely uncurated internet data, which may contain biased or exclusionary language, potentially influencing the model's outputs in unintended ways.

\section{Conclusion}

In this work, we propose a pipeline for in-context emotion recognition that highlights the importance of simplicity and the use of high-level semantic information in the task. Our approach effectively mitigates noise from raw visual inputs by generating context descriptions in natural language, leveraging the vast common sense reasoning of pre-trained VLLMs for each sample and fusing them with visual features. 
Our method achieves state-of-the-art results on one video and two image datasets, namely BoLD, EMOTIC, and CAER-S, demonstrating the simplicity and effectiveness of VLLMs-based context generation over intricate architectural designs.


{
    \small
    \begin{thebibliography}{10}\itemsep=-1pt

\bibitem{hci}
A.~A. Alnuaim, M.~Zakariah, A.~Alhadlaq, C.~Shashidhar, W.~A. Hatamleh, H.~Tarazi, P.~K. Shukla, and R.~Ratna.
\newblock Human-computer interaction with detection of speaker emotions using convolution neural networks.
\newblock {\em Computational Intelligence and Neuroscience}, 2024.

\bibitem{barrett_context_2011}
L.~F. Barrett, B.~Mesquita, and M.~Gendron.
\newblock Context in {Emotion} {Perception}.
\newblock {\em Current Directions in Psychological Science}, 2011.

\bibitem{bisk2020experience}
Y.~Bisk, A.~Holtzman, J.~Thomason, J.~Andreas, Y.~Bengio, J.~Chai, M.~Lapata, A.~Lazaridou, J.~May, A.~Nisnevich, N.~Pinto, and J.~Turian.
\newblock Experience grounds language.
\newblock In B.~Webber, T.~Cohn, Y.~He, and Y.~Liu, editors, {\em Proceedings of the 2020 Conference on Empirical Methods in Natural Language Processing (EMNLP)}, pages 8718--8735. Association for Computational Linguistics, Nov. 2020.

\bibitem{cheng2024emotionllamamultimodalemotionrecognition}
Z.~Cheng, Z.-Q. Cheng, J.-Y. He, J.~Sun, K.~Wang, Y.~Lin, Z.~Lian, X.~Peng, and A.~Hauptmann.
\newblock Emotion-llama: Multimodal emotion recognition and reasoning with instruction tuning.
\newblock In {\em arXiv}, 2024.

\bibitem{robotics}
H.~Christensen, N.~Amato, H.~Yanco, M.~Mataric, H.~Choset, A.~Drobnis, K.~Goldberg, J.~Grizzle, G.~Hager, J.~Hollerbach, S.~Hutchinson, V.~Krovi, W.~S. D~Lee, and J.~Trinkle.
\newblock A roadmap for us robotics–from internet to robotics 2020 edition.
\newblock {\em Foundations and Trends in Robotics}, 2021.

\bibitem{dai2023instructblip}
W.~Dai, J.~Li, D.~Li, A.~M.~H. Tiong, J.~Zhao, W.~Wang, B.~Li, P.~Fung, and S.~Hoi.
\newblock Instructblip: Towards general-purpose vision-language models with instruction tuning.
\newblock In {\em arXiv}, 2023.

\bibitem{de_Lima_Costa_2023_CVPR}
W.~de~Lima~Costa, E.~Talavera, L.~S. Figueiredo, and V.~Teichrieb.
\newblock High-level context representation for emotion recognition in images.
\newblock In {\em Proceedings of the IEEE/CVF Conference on Computer Vision and Pattern Recognition (CVPR) Workshops}, 2023.

\bibitem{dosovitskiy2020image}
A.~Dosovitskiy, L.~Beyer, A.~Kolesnikov, D.~Weissenborn, X.~Zhai, T.~Unterthiner, M.~Dehghani, M.~Minderer, G.~Heigold, S.~Gelly, et~al.
\newblock An image is worth 16x16 words: Transformers for image recognition at scale.
\newblock {\em arXiv}, 2020.

\bibitem{fei2007we}
L.~Fei-Fei, A.~Iyer, C.~Koch, and P.~Perona.
\newblock What do we perceive in a glance of a real-world scene?
\newblock {\em Journal of vision}, 7(1):10--10, 2007.

\bibitem{filntisis_emotion_2020}
P.~P. Filntisis, N.~Efthymiou, G.~Potamianos, and P.~Maragos.
\newblock Emotion {Understanding} in {Videos} {Through} {Body}, {Context}, and {Visual}-{Semantic} {Embedding} {Loss}.
\newblock In {\em arXiv}, 2020.

\bibitem{foteinopoulou_emoclip_2023}
N.~M. Foteinopoulou and I.~Patras.
\newblock {EmoCLIP}: {A} {Vision}-{Language} {Method} for {Zero}-{Shot} {Video} {Facial} {Expression} {Recognition}.
\newblock In {\em IEEE 18th International Conference on Automatic Face and Gesture Recognition (FG)}, 2024.

\bibitem{gao2021graph}
Q.~Gao, H.~Zeng, G.~Li, and T.~Tong.
\newblock Graph reasoning-based emotion recognition network.
\newblock {\em IEEE Access}, 2021.

\bibitem{greenaway_context_2018}
K.~H. Greenaway, E.~K. Kalokerinos, and L.~A. Williams.
\newblock Context is {Everything} (in {Emotion} {Research}).
\newblock {\em Social and Personality Psychology Compass}, 2018.

\bibitem{hoang2021context}
M.-H. Hoang, S.-H. Kim, H.-J. Yang, and G.-S. Lee.
\newblock Context-aware emotion recognition based on visual relationship detection.
\newblock {\em IEEE Access}, 2021.

\bibitem{hu2022lora}
E.~J. Hu, Y.~Shen, P.~Wallis, Z.~Allen-Zhu, Y.~Li, S.~Wang, L.~Wang, and W.~Chen.
\newblock Lo{RA}: Low-rank adaptation of large language models.
\newblock In {\em International Conference on Learning Representations}, 2022.

\bibitem{jaiswal2020attention}
S.~Jaiswal, S.~Misra, and G.~Nandi.
\newblock Attention-guided context-aware emotional state recognition.
\newblock In {\em 2020 IEEE 7th Uttar Pradesh Section International Conference on Electrical, Electronics and Computer Engineering (UPCON)}, 2020.

\bibitem{kosti2017emotic}
R.~Kosti, J.~M. Alvarez, A.~Recasens, and A.~Lapedriza.
\newblock Emotic: Emotions in context dataset.
\newblock In {\em Proceedings of the IEEE Conference on Computer Vision and Pattern Recognition Workshops}, 2017.

\bibitem{lee2019context}
J.~Lee, S.~Kim, S.~Kim, J.~Park, and K.~Sohn.
\newblock Context-aware emotion recognition networks.
\newblock In {\em Proceedings of the IEEE/CVF international conference on computer vision}, 2019.

\bibitem{lei2024instructercreformingemotionrecognition}
S.~Lei, G.~Dong, X.~Wang, K.~Wang, and S.~Wang.
\newblock Instructerc: Reforming emotion recognition in conversation with a retrieval multi-task llms framework.
\newblock In {\em arXiv}, 2024.

\bibitem{li2023cliper}
H.~Li, H.~Niu, Z.~Zhu, and F.~Zhao.
\newblock Cliper: A unified vision-language framework for in-the-wild facial expression recognition.
\newblock In {\em arXiv}, 2023.

\bibitem{li_blip-2_2023}
J.~Li, D.~Li, S.~Savarese, and S.~Hoi.
\newblock {BLIP}-2: {Bootstrapping} {Language}-{Image} {Pre}-training with {Frozen} {Image} {Encoders} and {Large} {Language} {Models}.
\newblock In {\em arXiv}, 2023.

\bibitem{li2021human}
W.~Li, X.~Dong, and Y.~Wang.
\newblock Human emotion recognition with relational region-level analysis.
\newblock {\em IEEE Transactions on Affective Computing}, 2021.

\bibitem{li2021sequential}
X.~Li, X.~Peng, and C.~Ding.
\newblock Sequential interactive biased network for context-aware emotion recognition.
\newblock In {\em 2021 IEEE International Joint Conference on Biometrics (IJCB)}. IEEE, 2021.

\bibitem{liu2023improvedllava}
H.~Liu, C.~Li, Y.~Li, and Y.~J. Lee.
\newblock Improved baselines with visual instruction tuning.
\newblock In {\em arXiv}, 2023.

\bibitem{liu2023llava}
H.~Liu, C.~Li, Q.~Wu, and Y.~J. Lee.
\newblock Visual instruction tuning.
\newblock In {\em Advances in Neural Information Processing Systems}, 2023.

\bibitem{liu2019roberta}
Y.~Liu, M.~Ott, N.~Goyal, J.~Du, M.~Joshi, D.~Chen, O.~Levy, M.~Lewis, L.~Zettlemoyer, and V.~Stoyanov.
\newblock Roberta: A robustly optimized bert pretraining approach.
\newblock {\em arXiv}, 2019.

\bibitem{loshchilov2018decoupled}
I.~Loshchilov and F.~Hutter.
\newblock Decoupled weight decay regularization.
\newblock In {\em International Conference on Learning Representations}, 2018.

\bibitem{luo_arbee_2020}
Y.~Luo, J.~Ye, J.~Adams, J.~Li, M.~G. Newman, and J.~Z. Wang.
\newblock {ARBEE}: {Towards} {Automated} {Recognition} of {Bodily} {Expression} of {Emotion} {In} the {Wild}.
\newblock {\em International Journal of Computer Vision}, 2020.

\bibitem{mittal_emoticon_2020}
T.~Mittal, P.~Guhan, U.~Bhattacharya, R.~Chandra, A.~Bera, and D.~Manocha.
\newblock {EmotiCon}: {Context}-{Aware} {Multimodal} {Emotion} {Recognition} {Using} {Frege}'s {Principle}.
\newblock In {\em Proceedings of the IEEE/CVF Conference on Computer Vision and Pattern Recognition}, 2020.

\bibitem{nicolaou2011}
M.~A. Nicolaou, H.~Gunes, and M.~Pantic.
\newblock Continuous prediction of spontaneous affect from multiple cues and modalities in valence-arousal space.
\newblock {\em IEEE Transactions on Affective Computing}, 2011.

\bibitem{pikoulis_leveraging_2021}
I.~Pikoulis, P.~P. Filntisis, and P.~Maragos.
\newblock Leveraging {Semantic} {Scene} {Characteristics} and {Multi}-{Stream} {Convolutional} {Architectures} in a {Contextual} {Approach} for {Video}-{Based} {Visual} {Emotion} {Recognition} in the {Wild}.
\newblock In {\em 2021 16th {IEEE} {International} {Conference} on {Automatic} {Face} and {Gesture} {Recognition} ({FG} 2021)}, 2021.

\bibitem{ruan2020iccv}
S.~Ruan, K.~Zhang, Y.~Wang, H.~Tao, W.~He, G.~Lv, and E.~Chen.
\newblock Context-aware generation-based net for multi-label visual emotion recognition.
\newblock In {\em IEEE International Conference on Multimedia and Expo (ICME)}, 2020.

\bibitem{singh2024llavafindsfreelunch}
S.~Singh, H.~S. I, Y.~Singla, V.~Baths, R.~R. Shah, C.~Chen, and B.~Krishnamurthy.
\newblock Llava finds free lunch: Teaching human behavior improves content understanding abilities of llms.
\newblock In {\em arXiv}, 2024.

\bibitem{education}
D.~Tanko, S.~Dogan, F.~B. Demir, M.~Baygin, S.~E. Sahin, and T.~Tuncer.
\newblock Shoelace pattern-based speech emotion recognition of the lecturers in distance education: Shoepat23.
\newblock {\em Applied Acoustics}, 2022.

\bibitem{TREISMAN1980}
A.~M. Treisman and G.~Gelade.
\newblock A feature-integration theory of attention.
\newblock {\em Cognitive Psychology}, 12(1):97--136, 1980.

\bibitem{vaswani2017attention}
A.~Vaswani, N.~Shazeer, N.~Parmar, J.~Uszkoreit, L.~Jones, A.~N. Gomez, L.~u. Kaiser, and I.~Polosukhin.
\newblock Attention is all you need.
\newblock In {\em Advances in Neural Information Processing Systems}, 2017.

\bibitem{Xie_2024_CVPR}
H.~Xie, C.-J. Peng, Y.-W. Tseng, H.-J. Chen, C.-F. Hsu, H.-H. Shuai, and W.-H. Cheng.
\newblock Emovit: Revolutionizing emotion insights with visual instruction tuning.
\newblock In {\em Proceedings of the IEEE/CVF Conference on Computer Vision and Pattern Recognition (CVPR)}, 2024.

\bibitem{xu2023lvlm}
P.~Xu, W.~Shao, K.~Zhang, P.~Gao, S.~Liu, M.~Lei, F.~Meng, S.~Huang, Y.~Qiao, and P.~Luo.
\newblock Lvlm-ehub: A comprehensive evaluation benchmark for large vision-language models.
\newblock In {\em arXiv}, 2023.

\bibitem{yang_context_2023}
D.~Yang, Z.~Chen, Y.~Wang, S.~Wang, M.~Li, S.~Liu, X.~Zhao, S.~Huang, Z.~Dong, P.~Zhai, and L.~Zhang.
\newblock Context {De}-{Confounded} {Emotion} {Recognition}.
\newblock In {\em Proceedings of the IEEE/CVF Conference on Computer Vision and Pattern Recognition}, 2023.

\bibitem{10.1145/3503161.3547754}
D.~Yang, S.~Huang, H.~Kuang, Y.~Du, and L.~Zhang.
\newblock Disentangled representation learning for multimodal emotion recognition.
\newblock In {\em Proceedings of the 30th ACM International Conference on Multimedia}, 2022.

\bibitem{yang2022emotion}
D.~Yang, S.~Huang, S.~Wang, Y.~Liu, P.~Zhai, L.~Su, M.~Li, and L.~Zhang.
\newblock Emotion recognition for multiple context awareness.
\newblock In {\em European Conference on Computer Vision}, 2022.

\bibitem{10.1145/3503161.3547755}
D.~Yang, H.~Kuang, S.~Huang, and L.~Zhang.
\newblock Learning modality-specific and -agnostic representations for asynchronous multimodal language sequences.
\newblock In {\em Proceedings of the 30th ACM International Conference on Multimedia}, 2022.

\bibitem{yang2024robust}
D.~Yang, K.~Yang, M.~Li, S.~Wang, S.~Wang, and L.~Zhang.
\newblock Robust emotion recognition in context debiasing.
\newblock In {\em Proceedings of the IEEE/CVF Conference on Computer Vision and Pattern Recognition}, 2024.

\bibitem{yang2024emollmmultimodalemotionalunderstanding}
Q.~Yang, M.~Ye, and B.~Du.
\newblock Emollm: Multimodal emotional understanding meets large language models.
\newblock In {\em arXiv}, 2024.

\bibitem{zhang2019contextaware}
M.~Zhang, Y.~Liang, and H.~Ma.
\newblock Context-aware affective graph reasoning for emotion recognition.
\newblock In {\em 2019 IEEE International Conference on Multimedia and Expo (ICME)}, 2019.

\bibitem{zhang2024visualpromptingllmsenhancing}
Q.~Zhang, Z.~Wang, D.~Zhang, W.~Niu, S.~Caldwell, T.~Gedeon, Y.~Liu, and Z.~Qin.
\newblock Visual prompting in llms for enhancing emotion recognition, 2024.

\bibitem{zhang2023learning}
S.~Zhang, Y.~Pan, and J.~Z. Wang.
\newblock Learning emotion representations from verbal and nonverbal communication.
\newblock In {\em Proceedings of the IEEE/CVF Conference on Computer Vision and Pattern Recognition}, 2023.

\bibitem{zhang2024dialoguellmcontextemotionknowledgetuned}
Y.~Zhang, M.~Wang, Y.~Wu, P.~Tiwari, Q.~Li, B.~Wang, and J.~Qin.
\newblock Dialoguellm: Context and emotion knowledge-tuned large language models for emotion recognition in conversations.
\newblock In {\em arXiv}, 2024.

\bibitem{zhao2021robust}
Z.~Zhao, Q.~Liu, and F.~Zhou.
\newblock Robust lightweight facial expression recognition network with label distribution training.
\newblock In {\em Proceedings of the AAAI conference on artificial intelligence}, 2021.

\bibitem{zhao2023prompting}
Z.~Zhao and I.~Patras.
\newblock Prompting visual-language models for dynamic facial expression recognition.
\newblock In {\em British Machine Vision Conference (BMVC)}, 2023.

\end{thebibliography}
}
\end{document}